\documentclass{article} %
\usepackage{iclr2024_conference,times}
\renewcommand{\headrulewidth}{0pt} %

\usepackage{amsmath,amsfonts,bm}

\def\eqref#1{equation~\ref{#1}}

\def\1{\bm{1}}

\DeclareMathAlphabet{\mathsfit}{\encodingdefault}{\sfdefault}{m}{sl}
\SetMathAlphabet{\mathsfit}{bold}{\encodingdefault}{\sfdefault}{bx}{n}

\usepackage{graphicx}
\usepackage{subcaption}
\usepackage{wrapfig}
\usepackage{bbm}

\usepackage{url}
\usepackage{booktabs} %
\usepackage{multirow}
\usepackage[hidelinks,colorlinks,linkcolor=blue,filecolor=blue,citecolor=magenta,urlcolor=blue]{hyperref}

\usepackage{algpseudocode}
\usepackage{paralist}
\usepackage{bbm}
\usepackage{booktabs, diagbox}
\usepackage{soul}
\usepackage{makecell}
\usepackage{color, colortbl}

\definecolor{Gray}{gray}{0.9}

\usepackage[capitalise]{cleveref}
\crefname{assumption}{Assumption}{Assumptions}
\crefname{proposition}{Proposition}{Proposition}
\crefname{corollary}{Corollary}{Corollaries}
\crefname{algorithm}{Alg.}{Algs.}
\crefname{theorem}{Theorem}{Theorems}
\crefname{figure}{Fig.}{Figs.}
\crefname{appendix}{App.}{Apps.}
\crefname{section}{Sec.}{Secs.}

\title{Extending Context Window of Large Language Models via Semantic Compression}

\author{Weizhi Fei$^\dag$~$^\ddag$ \& Xueyan Niu\thanks{Correspondence to: \texttt{niuxueyan3@huawei.com}.}~~$^\ddag$ \\
$^\dag$~Department of Mathematical Sciences, Tsinghua University, Beijing, China\\
$^\ddag$~Theory Lab, 2012 Labs, Huawei Technologies Co., Ltd.\\
\AND
Pingyi Zhou, Lu Hou, Bo Bai, Lei Deng, Wei Han \\
Huawei Technologies Co., Ltd.
}

\iclrfinalcopy %
\begin{document}

\maketitle
\vspace{-1em}
\begin{abstract}
Transformer-based Large Language Models (LLMs) often impose limitations on the length of the text input to ensure the generation of fluent and relevant responses. This constraint restricts their applicability in scenarios involving long texts. 
We propose a novel semantic compression method that enables generalization to texts that are 6-8 times longer, without incurring significant computational costs or requiring fine-tuning. Our proposed framework draws inspiration from source coding in information theory and employs a pre-trained model to reduce the semantic redundancy of long inputs before passing them to the LLMs for downstream tasks. 
Experimental results demonstrate that our method effectively extends the context window of LLMs across a range of tasks including question answering, summarization, few-shot learning, and information retrieval. Furthermore, the proposed semantic compression method exhibits consistent fluency in text generation while reducing the associated computational overhead.
\end{abstract}

\section{Introduction}
The recent successful release of large language models (LLMs) such as ChatGPT \citep{radford2019language} and LLaMA \citep{touvron2023llama} has sparked significant research efforts from both industry and academia. These LLMs have demonstrated the ability to engage in fluent and coherent conversations with human users, and have shown exceptional performance across various tasks, including document summarization, question-answering, dialogue bots, and code generation copilots.

One critical issue faced by state-of-the-art (SoTA) LLMs is the restriction on the length of text that can be inputted into the model at once. When the input context exceeds the limit of the context window, the performance of these models rapidly declines. This limitation poses a challenge when it comes to handling long texts such as scientific papers, novels, and legal contracts with current LLMs. As a result, there has been a growing interest in finding ways to extend the input length without significantly compromising the model's performance.

The limitation on the context window primarily stems from the quadratic computation of the self-attention mechanism in the transformer. Handling lengthy texts significantly increases the computational costs in terms of memory and time. Typically, models are trained on short contexts, and the maximum sequence length (i.e., the context window) is determined. If the models are compelled to generate contexts that exceed the context window, they tend to compromise the quality of the output due to the lack of position encoding information during the training process. Furthermore, generating long sequences imposes substantial memory requirements on the computational device. This accumulation of memory requirements and the lack of effective position encoding can result in length generalization failure \citep{anil2022exploring}, where the models struggle to generate meaningful and coherent text beyond a certain context window size.

Some approaches have been developed to address the aforementioned challenges. One approach is to devise architectures with nearly linear complexity, which enables efficient scaling to handle very long sequences. However, training a large model from scratch incurs substantial cost. Another strategy involves employing interpolation and fine-tuning techniques to adapt the position encoding to unseen sequence lengths. While this method has the potential to compromise the overall performance of LLMs, it still demands significant time and GPU resources for fine-tuning and inference on long sequences. Therefore, it is more efficient and resource-friendly to design methods that do not necessitate altering the parameters of the pre-trained model.

While most previous algorithms relied on modifying the pre-trained model, we instead exploit the statistical properties of input natural language. One empirical phenomenon, known as Zipf's law \citep{zipf2016human}, observes that a small set of the most frequent word tokens in a large corpus of natural language account for almost all occurrences. This pattern arises from the tendency of language users to minimize effort in their daily conversations. Consequently, by utilizing an expanded vocabulary, sentences can be significantly shortened while preserving the same semantic meaning. Moreover, it is common for language users to include redundant words during communication \citep{strunk2007elements}. These language habits are prevalent among users, and we propose to include a semantic compression module to mitigate the redundancy associated with these habits.

\begin{figure}[tb]
\centering
    \includegraphics[width=.65\textwidth]{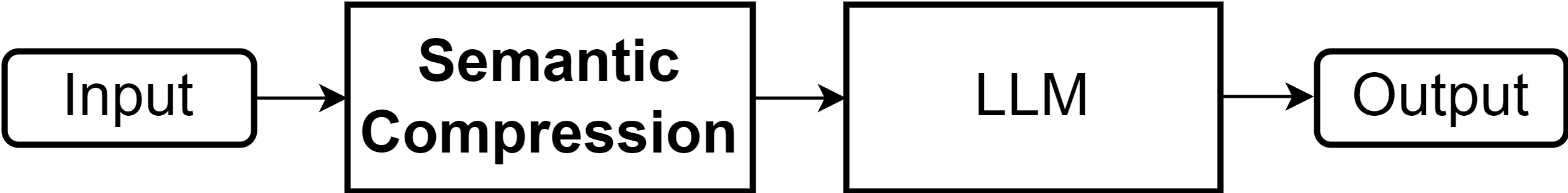}   
    \caption{\small With the inclusion of the semantic compression module, the redundancies in the input are eliminated, thereby effectively extending the context window. The semantic compression is reminiscent of the concept of source coding in information theory.}
    \label{fig:framework}
\end{figure}

Our proposed semantic compression method, reminiscent of lossy source coding in information theory, extends the context window by equivalently shortening the long text while preserving the semantic meaning. This procedure is conducted before inputting the tokens into the pre-trained LLMs.
As illustrated in Fig.~\ref{fig:framework}, the input undergoes compression before being transmitted to the LLM for various potential tasks. The semantic compression method can be customized and optimized for downstream tasks, taking into consideration practical constraints such as time and memory resources.
The implementation of the semantic compression module is straightforward and can easily be incorporated into other interpolation-based context window extension methods and black box APIs. It demonstrates enhanced performance compared to SoTA interpolation-based methods on a range of tasks, including single-document question answering, multi-document question answering, summarization, few-shot learning, and information retrieval, using real-world datasets while incurring no extra parameter updates or memory consumption. Empirically, the proposed method is computational efficient and achieves 6-8 times context window extension.

\paragraph{Our contributions:}
\begin{itemize}
    \item We introduce a context window extension framework for LLMs that utilizes semantic compression. This framework serves as a plug-and-play tool to mitigate redundancy in input texts by efficiently performing topic modeling. 
    \item  We construct a graph representation of the input to identify distinct sections of the text that pertain to different topics. The result is the segmentation of long texts into separate chunks, each focusing on a specific topic. We then conquer each chunk independently, resulting in a concise version of the original texts. This compression technique helps to condense the information while preserving the key ideas and context.
    \item We demonstrate the applicability of our proposed semantic compression method through extensive experiments. The results highlight the advantages of our method in several key applications, including single-document question answering, multi-document question answering, summarization, few-shot learning, and information retrieval.
\end{itemize}

\section{Related work}
With the advancement of SoTA LLMs, significant progress has been made in extending the context window lengths.

\subsection{Extrapolation and Interpolation} 
The mainstream line of research aims to adapt existing language models trained on short texts to accommodate longer ones during inference \citep{anil2022exploring}. The key idea is to modify the positional embedding, which has only been trained on short texts. Several studies are based on the Rotary Position Embeddings (RoPE) of LLaMA and methods of adjusting it to the longer sequences. \citet{chen2023extending} develops the Position Interpolation (PI) method to linearly scale the input positional indices. \citet{peng2023yarn} presents YaRN, an efficient extrapolate mechanism inspired by the neural tangent kernel, to extend the context window to $64$k and $128$k.

\subsection{Efficient Attention Operations}
Due to the self-attention mechanism, the inference cost of LLMs grows quadratically with the sequence length. Many methods have been proposed to decrease the complexity. \citet{dai-etal-2019-transformer} present Transformer-XL  which utilize segment-level recurrence agency and a novel positional encoding scheme. \citet{Beltagy2020Longformer} introduce Longformer with a sparse attention mechanism that scales linearly with sequence length. \citet{peng_bo_2021_5196578} provides a faster transformer, RWKV, which combines the strength of RNN and has linear complexity during inference. \citet{dao2022flashattention} propose FlashAttention, a chunking strategy for the input, and utilize recomputation to avoid the quadratic complexity of attention computation.  While these methods have the potential to handle longer input sequences \citep{ding2023longnet}, training new models can be costly. Moreover, these methods are not effective when dealing with out-of-distribution content lengths. 

The introduction of new positional embeddings requires fine-tuning on long sequences to adapt to the increased length, which can be computationally expensive. To address this, LongLoRA is introduced by \citet{longlora}, offering an efficient fine-tuning method with limited computational costs. More details on several other chunking strategies are provided in the survey by \citet{huang2023advancing}.

\subsection{Prompting}
There are ongoing efforts to extend the context window through smart prompting designs. \citet{wingate-etal-2022-prompt} utilize soft prompts to encode more information using fewer tokens. \citet{Chevalier2023AdaptingLM} present AutoCompressor, which utilizes soft prompts to compress the input sequence and then extends the original length of the base model.
Both \citet{zhou2023recurrentgpt} and \citet{wang2023recursively} recurrently apply LLMs to summarize the input texts to maintain long short-term memory for specific purposes such as story writing and dialogue generation, respectively.

\section{Methodology}
\begin{figure}[tb]
  \centering
  \includegraphics[width=.95\textwidth]{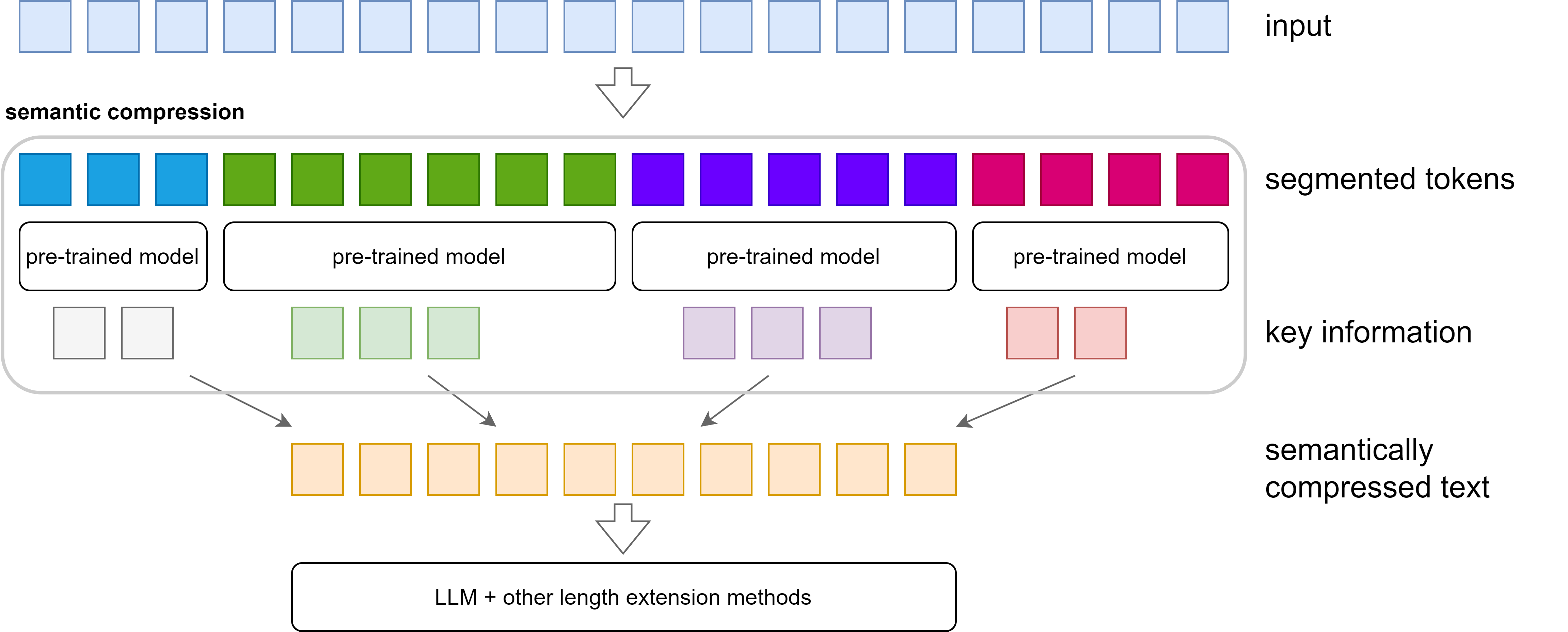}
  \caption{\small An illustration of our semantic compression method. The input text is initially segmented into topic-based chunks, utilizing the graph representation. Subsequently, these chunks undergo refinement using pre-trained models to ensure the preservation of key information. Finally, the refined chunks are assembled in accordance with the original order. The resulting texts, which have been semantically compressed, are approximately 6-8 times shorter in length compared to the original input. Consequently, they fall within the context window of the LLMs. Furthermore, for additional extension of the length, other methods such as extrapolation and interpolation-based techniques can be concatenated.
    }
  \label{fig:method}
\end{figure}

We propose our semantic compression method for extending the context window. The core idea is to compress the input into shorter texts without losing the key information and important details. This enables us to effectively include more content within the fixed input length constraint of the LLM.
Fig.~\ref{fig:method} provides an overview of our method, which leverages pre-trained summarization models commonly used in Natural Language Processing (NLP).

Existing summarization methods also have limitations regarding the length of the input. Here, we propose a divide-and-conquer based approach that takes into account the structure of the text. By identifying the topic structure of lengthy texts and dividing them into blocks that exhibit a certain level of mutual independence, the content within each block can be compressed efficiently due to their statistical correlation. Each block is then processed in parallel using pre-trained models, and the results are combined to create a condensed textual input that can be processed by the LLM. This approach aims to provide a more efficient and effective way of summarizing long texts by leveraging both the structure and content of the original text.

\subsection{Model}

Real-world textual content, such as speech and book, frequently displays hierarchical structures, wherein each section is structured around a particular topic, and different sections differ in topic in a sequential manner. This hierarchical structure, based on topics, bears resemblance to cliques in graphs. To identify this structure within long texts, we utilize weighted graphs to represent them and employ clustering methods to detect cliques in these graphs. The cliques can then be utilized to represent the topic-based content of the text, allowing us to obtain chunks based on the semantic relevance of the topics.

We begin by sequentially constructing sentence-level blocks within given lengths and representing them as nodes in our graph. In this step, we parse the text into different sentences or sub-sentences based on punctuation marks. Next, we sequentially fill the sentence-level blocks until they exceed the desired length before proceeding to the next blocks. Once we have obtained the sentence-level blocks, we connect the graph representation of long text $\mathcal{G}$ based on a pre-trained sentence embedding model (e.g., MiniLM \citep{wang2020minilm}), where the weight $\mathcal{G}[i][j]$ represents the semantic similarity between the $i$-th and $j$-th sentence-level blocks. Typically, this similarity is computed using cosine similarity, which measures the cosine of the angle between two embeddings. If the similarity between two blocks is higher, it indicates that they are closer in topics.

\subsection{Topic-Based Chunking}
We then apply clustering algorithms on the graph to identify the underlying topic structure. Within each cluster, we group the sentence-level blocks sequentially to obtain the topic-based chunks, which can then be handled simultaneously by the pre-trained model chosen according to the downstream task. The number of clusters can be adjusted to regulate the length of the text following semantic compression. If these semantic chunks still surpass the predetermined length, the identical procedure is repeated to acquire sub-level topic structures.

The obtained topic structures are tree-like, which can be flattened in accordance with the order of the original content. As per the model, each chunk is semantically centered around a specific topic, and these topics are mutually exclusive. Consequently, these chunks can be compressed independently by utilizing a pre-trained summarization model. Choosing from different pre-trained summarization models allows a trade-off between efficiency and effectiveness.  Consequently, we can opt to selectively substitute the original chunks with the output of these pre-trained models to ensure the preservation of the underlying topic structure. The semantic compressed text can be forwarded to the LLM directly or in combination with other extension schemes to further enhance the overall outcome.

\section{Experiments}
We demonstrate that the proposed method of semantic compression can effectively extend the context window by up to 7-8 times without modifying the parameters of the pre-trained models. 
Furthermore, the semantic compression module can be seamlessly integrated with existing methods, allowing for further extension of the context window. This versatility enables our approach to be adapted and combined with other techniques, enhancing the overall performance and flexibility.
To evaluate the performance of our method, we conduct experiments on several language tasks that require understanding of long contexts. These tasks include passkey retrieval, single-document question answering, multi-document question answering, summarization, and few-shot learning. In each task, the model is provided with a sequence of context $C$ (typically lengthy texts) and a sequence of text $Q$ (e.g., a prompt), and it is expected to generate the output answer $A$.
Additionally, we also investigate the perplexity metric \citep{peng2023yarn}, which measures the model's ability to predict the text and serves as an indicator of the fluency of the generated output. This analysis allows us to assess not only the effectiveness but also the quality of the generated output.

\subsection{Tasks and Datasets}
We begin by evaluating the proposed semantic compression method on various standard benchmark tasks, utilizing the pre-trained 7B LLaMA model \citep{touvron2023llama}. The original context window size of this model is $4096$. The tasks and datasets employed in our evaluation are sourced from the SCROLLS benchmark \citep{shaham2022scrolls} and LongBench \citep{bai2023longbench}. These datasets provide comprehensive and diverse contexts for our analysis.

\begin{figure}[t]
  \centering
  \includegraphics[width=.9\textwidth]{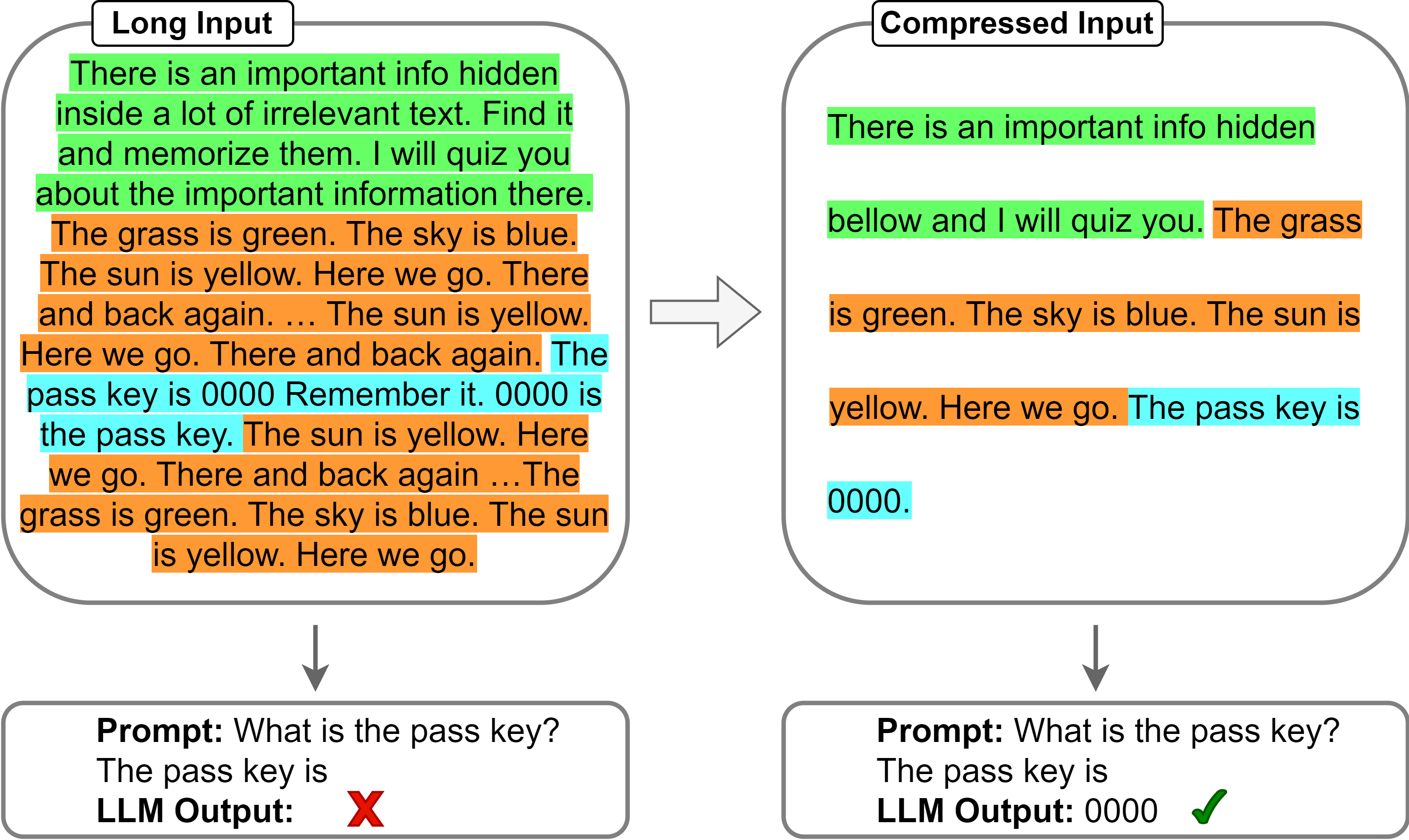}
  \caption{\small Example of synthetic prompt for the passkey retrieval task \citep{mohtashami2023landmark}. The pre-trained LLM is incapable of processing long input due to the context length constraint. By applying semantic compression, the redundant information in the long document is removed, and the compressed input retains essential key information. The LLM can then process the compressed input along with the prompt to generate the accurate answer. Notably, the distinct colors used in the illustration correspond to topic-based chunks.
    }
  \label{fig:passkey}
\end{figure}

\paragraph{Passkey Retrieval}
Retrieval has been an important application of LLMs. We evaluate the proposed method using a synthetic task for passkey retrieval introduced by \citet{mohtashami2023landmark}, where prompts are synthesized to conceal a generated passkey within a randomly chosen section of a long document. The passkey retrieval task assesses the model's capacity to extract important information from any position within lengthy contexts. An illustration of the task is shown in Fig.~\ref{fig:passkey}. The synthetic long text incorporates the passkey digits, and the task for the LLM is to retrieve these digits from the input text. 
Further specifics can be found in Appendix~\ref{sec:app:data}.

\paragraph{General NLP Tasks}
LongBench~\citep{bai2023longbench} is a multi-task benchmark designed for long text scenarios, consisting of six distinct tasks. In this study, we focus on the three English tasks from the set of four natural language tasks, namely single-document question answering, multi-document question answering, summarization, and few-shot learning. Each of the selected datasets contains 200 instances. Further information can be found in Appendix~\ref{sec:app:data}.

\paragraph{Fluency}
We evaluate the fluency of our semantic compression method using the perplexity score, which is defined as the exponential of the average negative log-likelihood of the probabilistic model $P$ on the distribution $D,$ i.e., 
\[
\mathrm{PPL}(D,P):= \exp (-\mathbb{E}_{x\in D} \log P(x)).
\]
A smaller perplexity score indicates more fluent sequences that are consistent with the model.

\subsection{Baselines}
We choose SoTA solutions from each mainstream approach as our baselines.

\paragraph{Fixed-size chunking}
To accommodate long context within a fixed-size context window, chunking is a straightforward yet efficient approach. In NLP related applications, large pieces of text are usually broken down into smaller segments for targeted applications. When the input length exceeds the context window, the fixed-size chunking method \citep{bai2023longbench} truncates the input sequence from the middle. This is because the most significant information typically resides at the beginning and end of the sequence.

\paragraph{Interpolation-based method}
YaRN~\citep{peng2023yarn} is a computationally efficient method for interpolating position encoding, which dynamically adjusts the Relative Positional Encoding (RoPE) over dimensions and scales the attention. YaRN offers multiple length-extended models for different versions of Llama2, with the models being trained on a total of 64 GPUs from 8 $\times$ A100 machines. In order to ensure a fair comparison, we choose the model based on Llama2 7B, adjusted from 4k to 64k, as our baseline.

\paragraph{Fine-tuning approach}
LongLoRA~\citep{longlora} is an efficient approach for fine-tuning that combines LoRA and shifts sparse attention to reduce computational costs. LongLoRA applies this technique to Llama2 models of different sizes, ranging from Llama2 7B, Llama2 13B, to Llama2 70B, with token lengths extended from 4k to 32k on a single 8 $\times \text{A100}$ device. In order to ensure a fair and unbiased comparison, we choose the Llama2 7B model with context extension achieved through improved LoRA fine-tuning as our baseline.

\section{Results}
We report the main results along with a comprehensive analysis.

\begin{figure}[ht]
  \centering
  \begin{minipage}[t]{0.48\linewidth}
    \centering
    \includegraphics[width=\linewidth]{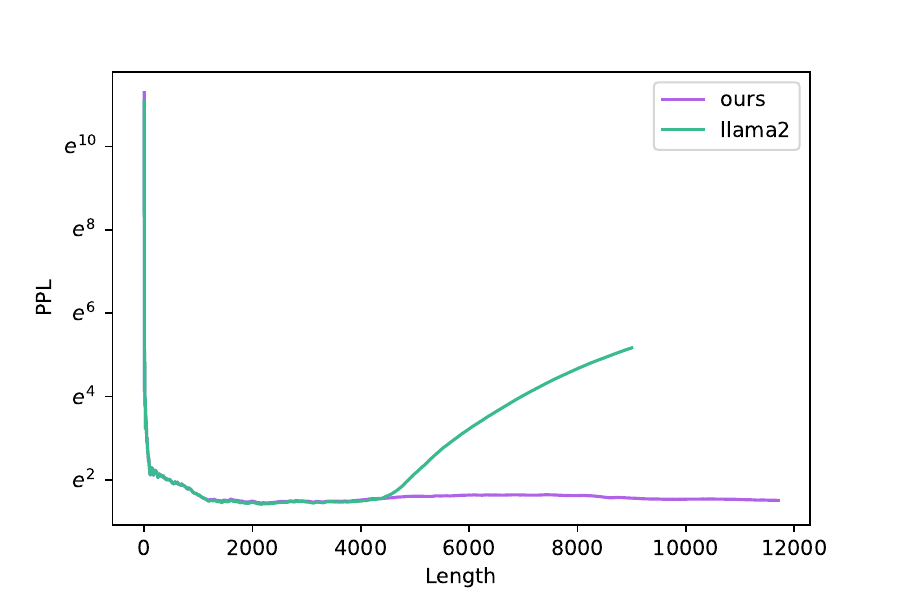}
    \caption{\small Perplexity on the GovReport dataset was evaluated at different sequence lengths. The perplexity curves of Llama2 (green) and our method (purple) exhibit similar trends for sequences up to 4k in length. However, as the sequence length exceeds the training length of 4k, our method effectively flattens the perplexity curve, indicating that fluency is preserved for longer sequences.}
    \label{fig:ppl_result}
  \end{minipage}
  \hfill
  \begin{minipage}[t]{0.48\linewidth}
    \centering
    \includegraphics[width=\linewidth]{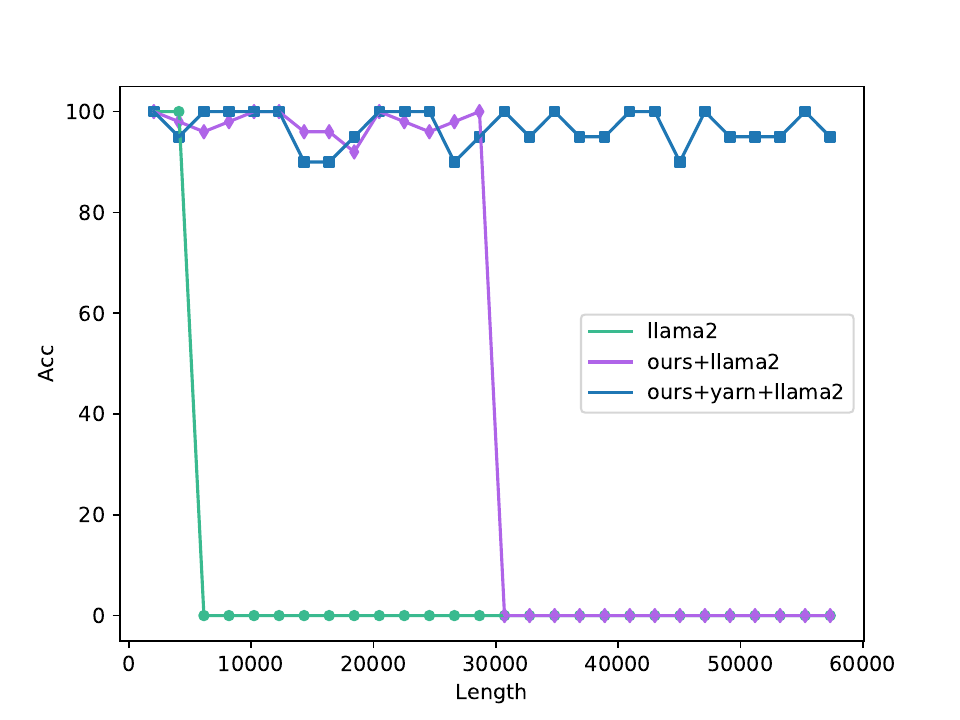}
    \caption{\small Comparison between model variants on the passkey retrieval task. The retrieval accuracy of the Llama2 baseline (green) drops to zero at about 5k due to out-of-memory issues. Our method (purple) successfully extends the length to 30k. Moreover, when combined with SoTA extrapolation-based method YaRN, the context length can be further extended to over 60k ensuring that the retrieval accuracy remains consistently above 90\%.}
    \label{fig:passkey_result}
  \end{minipage}
\end{figure}

\paragraph{Fluency}
We utilize the Llama2 model as our baseline to evaluate the fluency of generated texts by calculating the perplexity (PPL) score. Samples from the GovReport dataset are selected at varying lengths, and the reference texts are compared to the generated texts during the computation. In cases where the length of the input text exceeds the context window of Llama2, our semantic compression module shortens the input, thereby allowing the model to continue generating new content fluently. The resulting scores are depicted in Fig.~\ref{fig:ppl_result}.
The plots indicate that the perplexity of Llama2 initially decreases, but once it surpasses the window length, it rapidly increases. However, when our semantic compression method is employed, the PPL remains consistently low. This suggests that our approach successfully extends the context window up to three times without compromising the generation quality of the language model.

\paragraph{Passkey Retrieval}
We present the results of the passkey retrieval task in Fig.~\ref{fig:passkey_result}. When employing Llama2 for passkey retrieval, we observe a rapid drop in accuracy to zero once the input length surpasses the window size of $4096$. However, by utilizing our method, the retrieval accuracy of the Llama2 model remains above 90\% even for inputs with lengths of up to 30,000. This indicates that the semantic compression method extends the context window size of the language model by approximately 7-8 times. Furthermore, we combine our method with the SoTA interpolation-based method, YaRN, to further expand the context window size to up to 60,000, while consistently maintaining an accuracy above 90\%.

\paragraph{General NLP Tasks}
We present our results on various general NLP tasks in Table~\ref{tab:main}, including single-document question answering, multi-document question answering, summarization, and few-shot learning. When the token length is less than 4k, there is no need to compress the context, and our method performs at the same level as the original Llama2 model. However, both the interpolation-based method YaRN and the fine-tuning approach LongLora negatively impact the performance of the Llama2 model across almost all tasks. In the 4k-8k range, our method outperforms others in 8 out of 11 tasks. It is worth noting that our model performs slightly worse in the few-shot learning task. This can be attributed to the fact that few-shot learning necessitates more detailed information, whereas our compression scheme maintains information within a fixed window. Moving on to the 8k-16k range, our method achieves the best results in 9 out of 12 tasks, exhibiting similar performance to the 4k-8k range. In the 16k-32k range, our method outperforms others in 6 out of 11 tasks.
In the 32k+ range, other methods fail due to out-of-memory issues, while our method still maintains 70\% of the performance achieved in the 4k range.

\begin{table*}
    \centering
    \setlength{\tabcolsep}{3pt}

\begin{tabular}{l|l|p{0.9cm}p{1.4cm}|p{0.8cm}p{0.8cm}|p{0.8cm}p{0.8cm}|p{0.7cm}}
  \toprule
  Task & \diagbox{Dataset\\(length)}{Method} & \thead{Long\\LoRA} & \thead{Long\\LoRA (4k)} & yarn & yarn (4k) &  ours  & ours (4k) & 4k  \\
  \midrule
  & \textbf{4k-8k}   \\
  \midrule
  \multirow{3}{*}{Single-Doc QA} 
  & NarrativeQA & - & - & - & - & - & - & $18.7$\\
  & Qasper  & $11.6$ & $11.8$ & $13.4$ & $12.1$ & $\bm{23.4}$ &  $30.6$ & $19.2$\\
  & MultiFieldQA-en  & $24.5$ & $13.2$ & $34.9$ & $32.9$ & $\bm{37.4}$  & $58.7$ & $36.8$\\
  \midrule
  \multirow{3}{*}{Multi-Doc QA} 
  & HotpotQA & $11.5$ & $8.3$ & $11.3$ & $22.6$ & $\bm{50.6}$  & $50.0$ & $25.4$\\
  & 2WikiMultihopQA  & $10.1$ & $10.6$ & $8.9$ & $14.4$ & $\bm{29.8}$ & $61.8$ & $32.8$ \\
  & MuSiQue  & $10.0$ & - &$21.1$ & - & $\bm{50.0}$  & - & $9.4$\\
  \midrule
   \multirow{3}{*}{Summarization} 
  & GovReport & $24.7$ & $28.9$ &$28.8$ & $35.0$ & $\bm{31.8}$  & $32.2$ & $27.3$\\
  & QMSum  & $20.3$  & $17.0$ & $\bm{22.8}$  & $18.7$ & $21.1$ & $22.2$ & $20.8$\\
  & MutiNews  & $0.0$ & $0.0$ & $1.2$ & $18.9$ & $\bm{23.2}$  & $27.8$ & $25.8$\\
  \midrule
   \multirow{3}{*}{\makecell{Few-Shot\\Learning}} 
  & TREC & $65.8$  & $54.2$ & $\bm{70.9}$  & $50.0$ & $55.7$ & $54.2$ & $61.5$\\
  & TriviaQA  & $87.6$ & $80.6$ & $\bm{90.9}$ & $88.9$ & $83.3$  & $75.0$ & $77.8$\\
  & SAMSum  & $\bm{43.1}$  & $40.8$ & $40.4$  & $39.9$ & $41.6$  & $43.3$ & $40.7$\\
  \midrule
  & \textbf{8k-16k}   \\
  \midrule
  \multirow{3}{*}{Single-Doc QA} 
  & NarrativeQA &  $9.2$ & - & $13.9$ & - & $\bm{19.6}$ & - & $18.7$\\
  & Qasper  & - & 11.8 & $10.3$ & $12.1$ & $\bm{20.9}$  & $30.1$ & $19.2$\\
  & MultiFieldQA-en  & $22.5$ & $13.2$ & $18.9$ & $32.9$ & $\bm{35.9}$  & $58.7$ & $36.8$\\
  \midrule
  \multirow{3}{*}{Multi-Doc QA} 
  & HotpotQA & $8.9$ & $8.3$ & $8.7$ & $22.6$ & $\bm{28.1}$  & $50.0$ & $25.4$\\
  & 2WikiMultihopQA  & $9.5$ &$10.6$ & $9.9$ &$14.4$ & $\bm{26.3}$  & $61.8$ & $32.8$\\
  & MuSiQue  & $6.1$ & - &$4.2$ & - & $\bm{16.8}$  & - & $9.4$\\
  \midrule
   \multirow{3}{*}{Summarization} 
  & GovReport & $24.0$  & $28.9$ & $25.1$  & $35.0$ & $\bm{27.3}$  & $32.2$ & $27.3$\\
  & QMSum  & $22.5$  & $17.0$ & $21.8$  & $18.7$ & $\bm{23.4}$ & $22.2$ & $20.8$\\
  & MutiNews  & $0.0$ & $0.0$ & $0.0$ & $18.9$ & $\bm{22.0}$  & $27.8$ & $25.8$\\
  \midrule
   \multirow{3}{*}{\makecell{Few-Shot\\Learning}} 
  & TREC & $\bm{80.4}$  & $54.2$ & $77.3$  & $50.0$  & $57.7$ & $54.2$ & $61.5$\\
  & TriviaQA  & $86.5$ & $80.6$ & $\bm{89.1}$ & $88.9$ & $78.7$  & $75.0$ & $77.8$\\
  & SAMSum  & $\bm{44.5}$ & $40.8$ & $43.8$ & $39.9$ & $41.7$  & $43.3$ & $40.7$\\
  \midrule
  & \textbf{16k-32k}   \\
  \midrule
  \multirow{3}{*}{Single-Doc QA} 
  & NarrativeQA & $\bm{12.4}$ & - & $8.6$ & - & $9.8$  & - & $18.7$\\
  & Qasper  &  - & 11.8 & $9.2$ & $12.1$ & $\bm{15.2}$  & $30.1$ & $19.2$\\
  & MultiFieldQA-en  & $\bm{36.5}$ & $13.2$ &  $\bm{32.6}$ & $32.9$ & $23.6$  & $58.7$ & $36.8$\\
  \midrule
  \multirow{3}{*}{Multi-Doc QA} 
  & HotpotQA & $9.3$ & $8.3$ & $10.1$ & $22.6$ & $\bm{25.7}$  & $50.0$ & $25.4$\\
  & 2WikiMultihopQA  & $7.9$ & $10.6$ &$10.7$ & $14.4$  & $\bm{30.4}$ & $61.8$ & $32.8$\\
  & MuSiQue  & $5.4$  & - & $5.0$  & - & $\bm{14.6}$  & - & $9.4$\\
  \midrule
   \multirow{3}{*}{Summarization} 
  & GovReport & $24.7$ & $28.9$ & $\bm{26.4}$ & $35.0$ & $25.4$ & $32.2$ & $27.3$\\
  & QMSum  & $20.0$ & $17.0$ & $20.8$ & $18.7$ & $\bm{21.2}$  & $22.2$ & $21.5$\\
  & MutiNews  & $0.3$ &$0.0$ & $0.3$ &$18.9$ & $\bm{21.1}$  & $27.8$ & $26.4$\\
  \midrule
   \multirow{3}{*}{\makecell{Few-Shot\\Learning}} 
  & TREC & -  & $54.2$ & -  & $50.0$ & -  & $54.2$ & $61.5$\\
  & TriviaQA  & $88.8$ & $80.6$ & $\bm{90.1}$ & $88.9$ & $81.1$  & $75.0$ & $77.8$\\
  & SAMSum  & $\bm{44.7}$ & $40.8$ & $43.6$ & $39.9$ & $39.4$  & $43.3$ & $40.7$\\
  \bottomrule
  & \textbf{32k+}   \\
  \midrule
  \multirow{1}{*}{Single-Doc QA} 
  & NarrativeQA & oom & - & oom & - & $19.0$  & - & $18.7$\\
  \midrule
   \multirow{2}{*}{Summarization} 
  & GovReport & oom & $28.9$ & oom & $35.0$ & $21.7$ & $32.2$ & $27.3$\\
  & QMSum  & oom & $17.0$ & oom & $18.7$ & $22.4$  & $22.2$ & $21.5$\\
  \bottomrule
\end{tabular}

    \caption{\small Comparison of our semantic compression method with other baseline methods on a variety of tasks from the LongBench dataset. Method (4k) denotes evaluation results on texts shorter than 4k. The last column, labeled 4k, showcases the performance of the Llama2-7B-chat-4k baseline. Notably, our method consistently outperforms or achieves similar results compared to other SoTA length extension methods.} \label{tab:main}
\end{table*}

\section{Conclusion}
In this work, we propose a novel approach to addressing the limitation of input length in large language models using semantic compression. By leveraging the statistical properties of natural language and exploiting redundancy in communication, we are able to significantly shorten texts while preserving their semantic meaning. This allows for a 6-8 time extension of the context window without the need for modifying the parameters of the pre-trained model or incurring additional computational costs. Furthermore, the implementation of our semantic compression module is straightforward and can be easily integrated into other interpolation-based methods and black box APIs. This provides flexibility and adaptability to different downstream tasks, considering practical constraints such as time and memory resources.
We believe our work can lead to simpler context window extension method to be used in practice, thereby reducing the cost of large language models.

\bibliography{ref}
\bibliographystyle{iclr2024_conference}

\appendix
\section{Datasets}\label{sec:app:data}
\paragraph{Single-Doc QA}
\begin{itemize}
    \item \textbf{NarrativeQA} \citep{kovcisky2018narrativeqa} is a standard question-answering dataset that includes books from Project Gutenberg3 and movie screenplays from a list of websites. Question-answer pairs were provided by annotators, so that each of the 1,567 books and scripts has about 30 questions and answers, and two reference answers are given for each question.
    \item \textbf{Qasper} \citep{dasigi2021dataset} is a question-answering dataset of NLP publications containing abstractive, extractive, and yes/no questions.
    \item \textbf{MultiFieldQA-en} \citep{bai2023longbench} is a dataset created from multiple sources including legal documents, government reports, encyclopedias, and academic publications. Doctoral students were requested to annotate each article's queries and responses.
\end{itemize}

\paragraph{Multi-Doc QA}
\begin{itemize}
    \item \textbf{HotpotQA} \citep{yang2018hotpotqa} includes many 2-hop questions written by native speakers based on two related paragraphs.
    \item \textbf{2WikiMultihopQA} \citep{ho-etal-2020-constructing} involves up-to 5-hop questions systematacially  constructed by manual templates. Answering these questions requires reasoning paths and can not be solved by local content.
    \item \textbf{MuSiQue}\citep{trivedi-etal-2022-musique} consists of up to 4-hop questions and removes shortcuts and naturalness questions. Each question contains 2-4 supplement paragraphs which present the reasoning path and related paragraphs.
\end{itemize}

\paragraph{Summarization}
\begin{itemize}
    \item \textbf{GovReport} \citep{huang-etal-2021-efficient} collects detailed reports containing human-written summaries from  the U.S. Government Accountability Office and Congressional Research Service. These reports span a wide variety of national policy issues. 
    \item \textbf{QMSum} \citep{zhong2021qmsum} contains annotated meeting-summary pairs across many domains including including product, academic, and committee meetings.
    \item \textbf{MultiNews}\citep{fabbri2019multi} is  a multi-document summarization dataset. \citep{bai2023longbench}  cluster  2-10 news articles discussing the same event or topic, each paired with a human-written summary and form a new long text summarization task.
\end{itemize}

\paragraph{Few-Shot Learning}
To construct few-shot learning with long text, \citep{bai2023longbench} select a range of training examples in the following datasets to concatenate the context in LongBench.
\begin{itemize}
    \item \textbf{TREC} \citep{li2002trec} is a classification dataset with fine-grained class label.
    \item \textbf{TriviaQA} \citep{zhong2021qmsum} is a classification dataset and involves messenger-like conversations with human-written summaries.
    \item \textbf{SAMSum} \citep{fabbri2019multi}  reading comprehension dataset and consists of question-answer pairs annotated with evidence passages.
\end{itemize}

\begin{figure}[t]
  \centering
  \includegraphics[width=.75\textwidth]{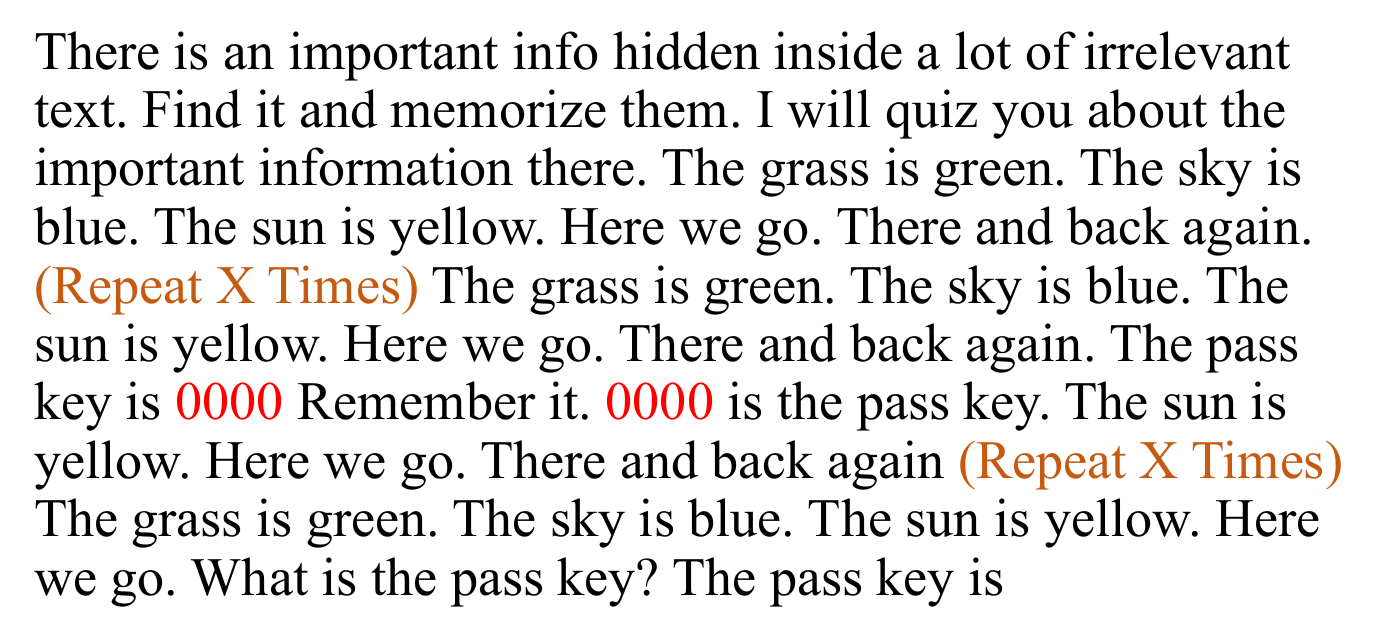}
  \caption{\small The query prompt contains task descriptions, redundant information, passkey information, redundant information, and query information. The passkey information is randomly placed
    within the text, while the remaining space up to a specified length is filled 
    with redundant information.
    }
  \label{fig:passkey-format}
\end{figure}

\paragraph{Passkey}
The randomly generated prompts of the passkey retrieval task is in the format of Fig.~\ref{fig:passkey-format}.

\section{Implementation Details}
In this section, we provide details of our algorithm implementation. Our algorithm utilizes several mature open-source models. For graph representation, we make use of the sentence similarity models all-MiniLM-L6-v2 provided by the Sentence Transformer platform, which can be found at the following link: \url{https://huggingface.co/sentence-transformers/all-MiniLM-L6-v2}.
For semantic compression, we employ the pre-trained model distilbart-cnn-12-6\footnote{Available at: \url{https://huggingface.co/sshleifer/distilbart-cnn-12-6}}. In most of our experiments, we utilize Llama2-7B-chat-4k as the base large language model~\citep{touvron2023llama}. The experiments were conducted on a single A40 GPU with 48GB memory.

\section{Complexity}

Given a context with length $L$, the origin complexity is $\mathcal{O}(L^2)$. Considering the length limitations of the compression module, we assume it has a minimum input length $\gamma_1$ and a maximum input length $\gamma_2$.
We  denote the compression ratio as $\alpha$.
Our method utilizes a divide-and-conquer strategy, dividing the long text into chunks where the total length is represented as $L = l_1  + \cdots + l_k$, and each chunk's length, $l_i$, satisfies the condition $\gamma_1 \le \l_i \le \gamma_2$. By $ k \gamma_1 \le L$, we can bound the complexity of the compression module 
\begin{equation}
    \sum_{i=1}^{k} l_i^2 \le \sum_{i=1}^{k} \gamma_2^2 = k\gamma_2^2 \le \frac{\gamma_2^2}{\gamma_1} L. 
\end{equation}
The complexity of inferring the compressed context is 
\begin{equation}
     (\sum_{i=1}^{k} \alpha l_i)^2 = (\alpha \sum_{i=1}^{k} l_i)^2 = \alpha^2 L^2. 
\end{equation}

Thus the main complexity of our algorithms can be bounded by $\frac{\gamma_2^2}{\gamma_1} L + \alpha^2 L^2$.

The result suggests that our algorithm can reduce the computational complexity by a factor of the square of the compression ratio during the inference stage. The compression module exhibits linear growth and can be processed in parallel.

\end{document}